\documentclass[11pt]{article}
\usepackage[utf8]{inputenc}
\usepackage[margin=1in]{geometry}
\usepackage{amsmath,amssymb}
\usepackage{graphicx}
\usepackage{booktabs}
\usepackage{hyperref}
\usepackage{natbib}
\usepackage{xcolor}
\usepackage{multirow}

\title{Benchmarking Open-Source PPG Foundation Models for Biological Age Prediction}

\author{
  N. Brag\\
  Independent Researcher
}

\date{March 2026}

\begin{document}
\maketitle


\bigskip

\begin{abstract}
A model trained on 212,231 subjects to predict vascular age from PPG (AI-PPG Age, trained on UK Biobank finger PPG) fails when applied to a different clinical population---its predictions collapse to a narrow range (38--67 years) regardless of the patient's true age. A general-purpose foundation model trained on no age-related objective at all achieves lower error on the same data. We document this reversal, explain its cause, and show that open-source PPG embeddings reliably encode biological aging information across acquisition contexts.

We benchmark three open-source PPG models---Pulse-PPG, PaPaGei-S, and AI-PPG Age---on 906 surgical patients from PulseDB (VitalDB clinical finger PPG, 125\,Hz, 10-second segments), using frozen embeddings with Ridge regression and 5-fold stratified cross-validation. Pulse-PPG achieves MAE\,=\,9.28 years (R$^2$\,=\,0.388), outperforming both the task-specific AI-PPG Age in linear probe mode (MAE\,=\,9.72) and the combination of heart rate variability with all demographic features (MAE\,=\,9.59). Fusing Pulse-PPG embeddings with demographics yields MAE\,=\,8.22 years (R$^2$\,=\,0.517, $r$\,=\,0.725). The age-adjusted PPG age gap correlates significantly with diastolic blood pressure ($r$\,=\,$-$0.188, $p$\,=\,1.2\,$\times$\,10$^{-8}$), consistent with Apple's finding that PPG morphology captures vascular aging beyond chronological age. The performance gap with Apple's PpgAge (MAE\,=\,2.43 years) is attributable to dataset scale ($\sim$236$\times$ fewer subjects) and population differences, not model architecture: our learning curve shows no plateau at 906 subjects. All code is publicly available.
\end{abstract}

\section{Introduction}

Biological age---a measure of physiological deterioration independent of chronological age---is a powerful predictor of mortality, cardiovascular disease, and age-related conditions \citep{jylhava2017biological}. Traditional biological age clocks rely on blood biomarkers \citep{levine2018epigenetic}, DNA methylation \citep{horvath2013dna}, or clinical assessments, limiting their use to periodic clinical visits.

The photoplethysmogram (PPG), a ubiquitous optical signal captured by pulse oximeters and consumer smartwatches, encodes information about arterial stiffness, cardiac output, and autonomic function---all of which change with aging \citep{elgendi2012use}. In October 2025, Apple introduced PpgAge, a deep learning model trained on 213,593 Apple Health Study participants that predicted chronological age with MAE\,=\,2.43 years from wrist PPG and demonstrated that the PPG age gap (predicted $-$ chronological age) predicted incident cardiovascular disease (HR\,=\,1.46) \citep{miller2025ppgage}.

PpgAge's reliance on proprietary data (Apple Health Study) and closed-source models, however, limits reproducibility and broader adoption. Two questions remain unanswered: (1) Can open-source PPG foundation models replicate this capability? (2) Does the PPG age signal persist in clinical finger PPG, beyond Apple Watch wrist PPG?

Several open-source PPG foundation models have recently emerged. \textbf{Pulse-PPG} \citep{saha2025pulseppg} is a ResNet-based model pre-trained via contrastive learning on wearable PPG data, while \textbf{PaPaGei} \citep{pillai2025papagei} uses a mixture-of-experts ResNet architecture pre-trained on 20 million PPG segments from public datasets. Both produce general-purpose PPG embeddings that can be used for downstream tasks via linear probing.

A critical question for clinical translation is whether foundation model embeddings capture aging information beyond what can be extracted from simple heart rate (HR) and heart rate variability (HRV) features. Apple reported that HR/HRV alone achieved MAE\,=\,6.1 years on their large dataset (using 30-day aggregated statistics from thousands of segments per subject), compared to 2.43 years for their full model. If foundation model embeddings merely encode HR/HRV, their added complexity would be unjustified.

In this work, we benchmark Pulse-PPG and PaPaGei-S for chronological age prediction on PulseDB \citep{wang2023pulsedb}, a public dataset of 5,361 subjects with clinical finger PPG from VitalDB surgical patients and MIMIC-III ICU patients. We evaluate frozen embeddings with Ridge regression, compare against HR/HRV and demographic baselines, and test whether the PPG age gap associates with cardiovascular risk factors, an analysis motivated by the key clinical finding from PpgAge.

\section{Methods}

\subsection{Dataset}

We use PPG segments from the VitalDB subset of PulseDB \citep{wang2023pulsedb}, comprising clinical finger PPG recordings from surgical patients at Seoul National University Hospital. PulseDB provides pre-segmented 10-second windows at 125\,Hz (1,250 samples per segment) with per-segment annotations including age, sex, height, weight, BMI, and intra-operative systolic and diastolic blood pressure (SBP, DBP).

From 943 downloaded subject files, 906 were successfully loaded (37 excluded: 29 due to file corruption or malformed h5py object references, 8 due to age $<$ 18). We restrict our analysis to the VitalDB subset, as MIMIC-III subjects in PulseDB lack the complete per-segment demographic annotations (height, weight, BMI, and blood pressure) required for our demographic feature experiments. We use up to 50 segments per subject, sampled evenly across available segments, yielding 41,104 total segments. The age distribution spans 18--92 years (mean 60.9, SD 15.6), with 56.2\% male subjects.

\subsection{Foundation Models}

\textbf{Pulse-PPG} \citep{saha2025pulseppg}: A 12-block ResNet1D (128 base filters, kernel size 11, max pooling) pre-trained via contrastive learning on wearable PPG from a 100-day field study with 120 participants. Expects 50\,Hz input; we resample from 125\,Hz using scipy's signal.resample. Produces 512-dimensional embeddings directly from the forward pass.

\textbf{PaPaGei-S} \citep{pillai2025papagei}: A ResNet1D with Mixture-of-Experts (18 blocks, 32 base filters, 3 experts) pre-trained on 20M PPG segments from public datasets using morphology-augmented contrastive learning. Operates at native 125\,Hz. Produces 512-dimensional embeddings.

\textbf{AI-PPG Age} \citep{nie2025aippgage}: A 1D CNN with Squeeze-and-Excitation blocks trained end-to-end on 212,231 UK Biobank subjects for direct vascular age regression using a distribution-aware loss function. Input is a single averaged beat waveform (resampled to 100 samples); output is a scalar age prediction. We evaluate this model in two modes: (1) \textit{zero-shot}, using direct age predictions without any adaptation; and (2) \textit{linear probe}, extracting 192-dimensional pre-dense layer activations via a forward hook on the final convolutional stage and training Ridge regression on these embeddings. Beat templates are extracted by detecting systolic peaks in each 10-second segment, isolating peak-to-peak waveforms, resampling each to 100 samples, and averaging across all valid beats in the segment.

All three models are used as frozen feature extractors---no fine-tuning is performed. Input segments for Pulse-PPG and PaPaGei-S are z-score normalized per segment before embedding extraction.

\subsection{HR/HRV Feature Extraction}

To contextualize foundation model performance, we extract traditional heart rate and heart rate variability features from each 10-second PPG segment. Systolic peaks are detected using scipy's \texttt{find\_peaks} with minimum inter-peak distance of 0.4\,s (corresponding to 150\,BPM maximum) and minimum prominence of 0.1. Inter-beat intervals (IBI) are computed from successive peak locations and filtered to the physiologically plausible range of 0.3--2.0\,s (30--200\,BPM). Segments with fewer than 4 detected peaks or fewer than 3 valid IBIs are excluded.

From valid segments, we extract 7 features: mean heart rate (HR), HR standard deviation, HR range, mean IBI, standard deviation of normal-to-normal intervals (SDNN), root mean square of successive differences (RMSSD), and the proportion of successive IBI differences exceeding 50\,ms (pNN50). All 41,104 segments (100\%) produced valid HR/HRV features.

\subsection{Age Prediction Pipeline}

For each feature set, we train a Ridge regression (sklearn RidgeCV, $\alpha \in \{0.01, 0.1, 1, 10, 100, 1000\}$) to predict chronological age. Features are standardized (zero mean, unit variance) before regression.

We evaluate the following feature configurations:
\begin{enumerate}
    \item \textbf{Baseline}: predict the training set mean age for all subjects
    \item \textbf{HR only}: mean heart rate (1 feature)
    \item \textbf{HR + HRV}: all 7 HR/HRV features
    \item \textbf{Demographics only}: sex, BMI, height, weight, SBP, DBP (6 features)
    \item \textbf{HR/HRV + Demographics}: 7 HR/HRV + 6 demographic features (13 features)
    \item \textbf{PPG only}: 512-dim embeddings from each foundation model
    \item \textbf{PPG + Demographics}: concatenation of embeddings and demographic features (518 features)
\end{enumerate}

\subsection{Evaluation Protocol}

We use 5-fold stratified cross-validation at the subject level. Subjects are stratified by age decade and assigned to folds via round-robin within each stratum, ensuring balanced age distributions across folds. All segments from a given subject appear in the same fold (no segment-level data leakage).

For each test fold, segment-level predictions are averaged per subject to produce a single age estimate. We report subject-level MAE, R$^2$, and Pearson correlation ($r$), along with per-decade MAE breakdown.

\subsection{Age Gap Association Analysis}

Following \citet{miller2025ppgage}, we compute the PPG age gap as predicted age minus chronological age for each subject using cross-validated predictions (each subject's prediction comes from a fold where it was held out). We test Pearson correlations between the age gap and cardiovascular markers (SBP, DBP, BMI), both raw and adjusted for chronological age via linear residualization. The age-adjusted partial correlation removes the confound that both the age gap and health markers correlate with chronological age.

Statistical significance is assessed at $\alpha = 0.05$; given three simultaneous tests, the Bonferroni-corrected threshold is $\alpha = 0.017$.

\section{Results}

\subsection{Age Prediction Performance}

Table~\ref{tab:main} presents the main results across all models and feature configurations.

\begin{table}[h]
\centering
\caption{Age prediction performance (5-fold stratified CV, 906 VitalDB subjects). MAE in years.}
\label{tab:main}
\begin{tabular}{lccc}
\toprule
\textbf{Model} & \textbf{MAE $\pm$ SD} & \textbf{R$^2$} & \textbf{$r$} \\
\midrule
Baseline (predict mean) & 11.91 & --- & --- \\
\midrule
\multicolumn{4}{l}{\textit{Traditional cardiac features}} \\
\quad HR only & 11.83 $\pm$ 0.21 & 0.015 & 0.122 \\
\quad HR + HRV (7 features) & 11.49 $\pm$ 0.15 & 0.072 & 0.283 \\
\midrule
\multicolumn{4}{l}{\textit{Demographics only}} \\
\quad Sex only & 11.81 $\pm$ 0.13 & 0.007 & 0.087 \\
\quad All demographics & 10.04 $\pm$ 0.38 & 0.270 & 0.519 \\
\midrule
\multicolumn{4}{l}{\textit{Combined traditional features}} \\
\quad HR/HRV + Demographics & 9.59 $\pm$ 0.30 & 0.329 & 0.574 \\
\midrule
\multicolumn{4}{l}{\textit{PPG foundation models (linear probe, 5-fold CV)}} \\
\quad PaPaGei-S & 9.80 $\pm$ 0.34 & 0.313 & 0.583 \\
\quad Pulse-PPG & 9.28 $\pm$ 0.44 & 0.388 & 0.645 \\
\quad AI-PPG Age (zero-shot) & 11.50 & 0.043 & 0.313 \\
\quad AI-PPG Age (linear probe) & 9.72 $\pm$ 0.37 & 0.310 & 0.563 \\
\midrule
\multicolumn{4}{l}{\textit{PPG + Demographics fusion}} \\
\quad PaPaGei-S + Demographics & 8.56 $\pm$ 0.22 & 0.477 & 0.696 \\
\quad \textbf{Pulse-PPG + Demographics} & \textbf{8.22 $\pm$ 0.25} & \textbf{0.517} & \textbf{0.725} \\
\quad AI-PPG Age + Demographics & 8.58 $\pm$ 0.25 & 0.440 & 0.664 \\
\bottomrule
\end{tabular}
\end{table}

Traditional HR/HRV features provide minimal age prediction capability from 10-second segments (MAE\,=\,11.49, barely improving over the baseline of 11.91). Demographics alone are moderately informative (MAE\,=\,10.04), with anthropometric measurements (height, weight) and blood pressure contributing the most signal. Combining HR/HRV with demographics yields MAE\,=\,9.59---the best achievable without deep learning.

Pulse-PPG embeddings alone (MAE\,=\,9.28) outperform the HR/HRV + Demographics combination (MAE\,=\,9.59), demonstrating that foundation model embeddings capture morphological aging information in the PPG waveform that goes beyond heart rate, heart rate variability, and body measurements combined. The fusion of Pulse-PPG with demographics achieves the best overall performance (MAE\,=\,8.22, R$^2$\,=\,0.517), with low fold-to-fold variance ($\pm$0.25 years).

The AI-PPG Age model (trained on UK Biobank healthy subjects) shows poor zero-shot transfer to clinical VitalDB data (MAE\,=\,11.50), with predictions compressed to a narrow range (38--67 years) compared to the true age range (18--92 years). This is consistent with a population shift: the UK Biobank model is calibrated for healthy ambulatory subjects, not surgical patients under anesthesia. With linear probe fine-tuning, AI-PPG Age (linear probe) achieves MAE\,=\,9.72, still inferior to Pulse-PPG and PaPaGei-S which were pre-trained on more diverse PPG data.

\begin{figure}[h]
\centering
\includegraphics[width=\textwidth]{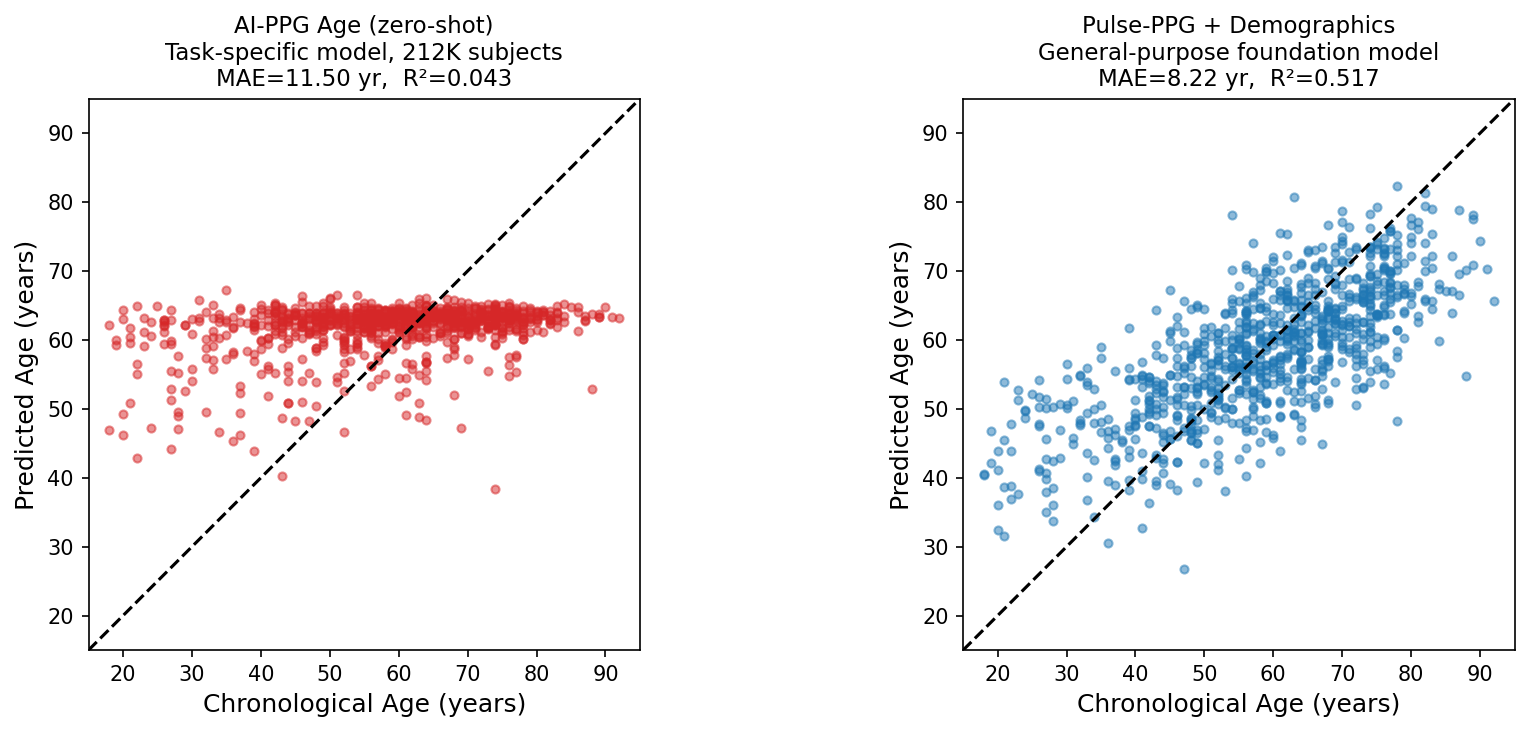}
\caption{Predicted vs.\ chronological age for the two contrasting models ($n$\,=\,906 subjects). \textit{Left}: AI-PPG Age (zero-shot), a task-specific model trained on 212,231 UK Biobank subjects, produces predictions compressed to a narrow range (38--67 years) regardless of true age --- a population shift failure. \textit{Right}: Pulse-PPG + Demographics, a general-purpose foundation model with no age-related training objective, achieves MAE\,=\,8.22 years with predictions well distributed along the diagonal.}
\label{fig:scatter}
\end{figure}

Figure~\ref{fig:scatter} shows predicted vs.\ chronological age for all 906 subjects under the best model, illustrating the regression-to-the-mean pattern at age extremes.

\subsection{Performance by Age Decade}

\begin{table}[h]
\centering
\caption{MAE by age decade for Pulse-PPG + Demographics ($n$\,=\,902 subjects aged 20--100; 4 subjects aged 18--19 excluded from decade bins).}
\label{tab:decade}
\begin{tabular}{lcccccc}
\toprule
\textbf{Age group} & 20--40 & 40--50 & 50--60 & 60--70 & 70--80 & 80--100 \\
\midrule
$n$ subjects & 96 & 140 & 213 & 229 & 179 & 45 \\
MAE (years) & 15.69 & 7.51 & 5.59 & 6.01 & 9.10 & 13.40 \\
\bottomrule
\end{tabular}
\end{table}

Performance is best for the 50--70 age range (MAE 5--6 years), which contains the majority of subjects. The model struggles with extremes ($<$40 and $>$80 years), consistent with regression toward the mean in an age-imbalanced dataset centered around 60 years.

\subsection{Learning Curve}

\begin{figure}[h]
\centering
\includegraphics[width=0.85\textwidth]{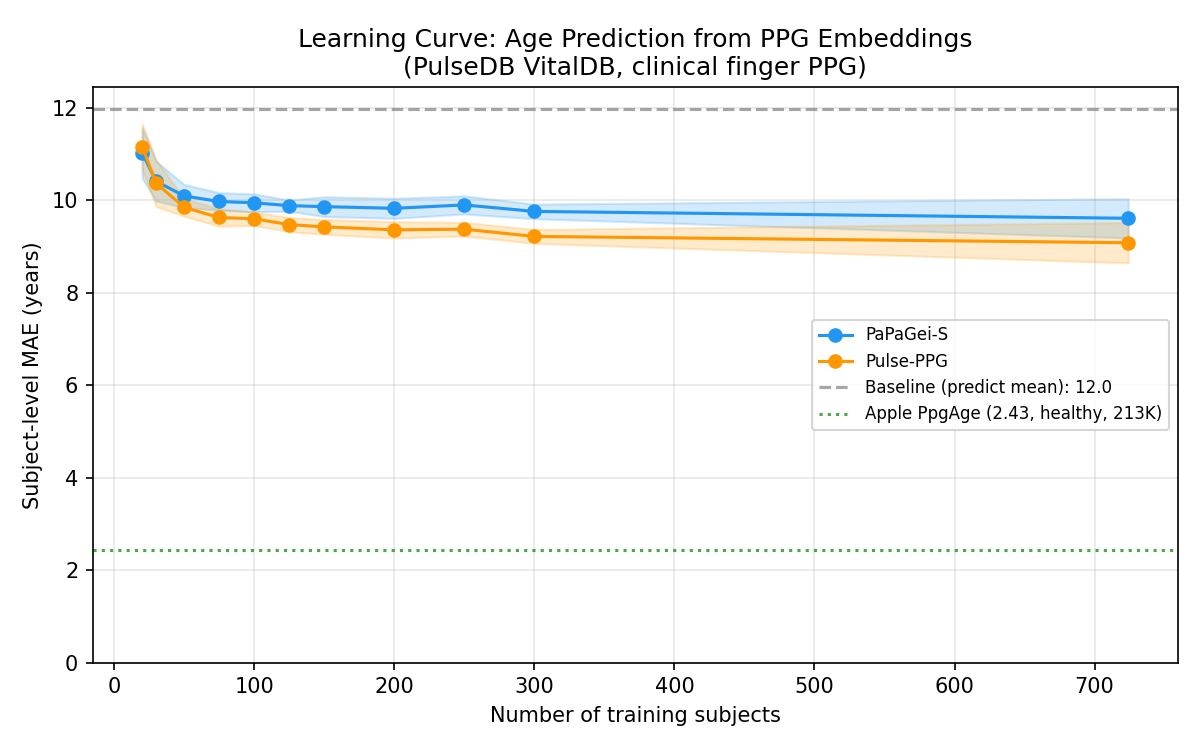}
\caption{Learning curve: subject-level MAE vs training set size for Pulse-PPG and PaPaGei-S (5-fold CV, averaged across folds).}
\label{fig:learning}
\end{figure}

Figure~\ref{fig:learning} shows that MAE decreases monotonically with training set size for both models, from $\sim$10.5 years (20 subjects) to $\sim$9.2 years (700+ subjects), without fully plateauing. This suggests that additional training data would further improve performance, and that the gap with Apple's PpgAge (MAE 2.43) is partly attributable to dataset size (906 vs 213,593 subjects).

\subsection{Age Gap Association with Cardiovascular Health}

Table~\ref{tab:gap} presents correlations between the cross-validated PPG age gap and cardiovascular markers.

\begin{table}[h]
\centering
\caption{Correlation of PPG age gap (Pulse-PPG + Demographics) with health markers ($n$\,=\,906).}
\label{tab:gap}
\begin{tabular}{lcccc}
\toprule
\textbf{Marker} & \textbf{$r$} & \textbf{$p$} & \textbf{$r_{\text{partial}}$} & \textbf{$p_{\text{partial}}$} \\
\midrule
SBP (mmHg) & $-$0.039 & 0.237 & 0.066 & 0.048 \\
DBP (mmHg) & 0.013 & 0.688 & $-$0.188 & \textbf{1.2\,$\times$\,10$^{-8}$} \\
BMI (kg/m$^2$) & $-$0.014 & 0.672 & $-$0.007 & 0.836 \\
\bottomrule
\end{tabular}
\end{table}

\begin{figure}[h]
\centering
\includegraphics[width=\textwidth]{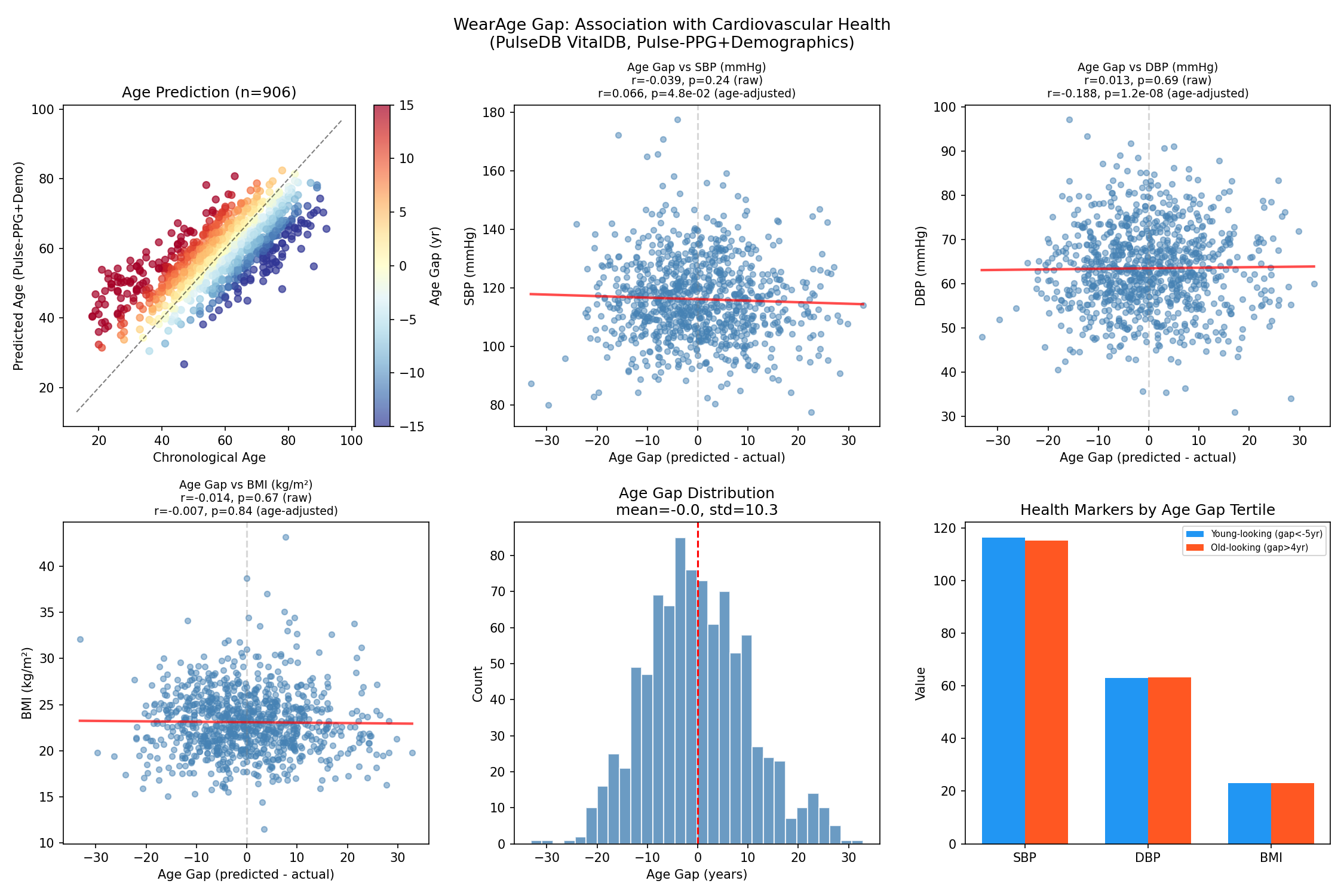}
\caption{Correlation of PPG age gap (Pulse-PPG + Demographics) with cardiovascular markers ($n$\,=\,906). Each panel shows the scatter plot with both raw and age-adjusted (partial) correlations. Only the DBP partial correlation survives Bonferroni correction.}
\label{fig:gap}
\end{figure}

The key finding is a highly significant age-adjusted partial correlation between the PPG age gap and DBP ($r$\,=\,$-$0.188, $p$\,=\,1.2\,$\times$\,10$^{-8}$). At a given chronological age, subjects whose PPG appears ``older'' have lower diastolic blood pressure. This is physiologically consistent: arterial stiffening with age reduces diastolic pressure (widening pulse pressure), and the PPG waveform directly reflects arterial compliance through its morphology \citep{millasseau2002contour}. A borderline association with SBP ($r$\,=\,0.066, $p$\,=\,0.048) is also observed, with the expected positive direction (stiffer arteries raise systolic pressure). Together, these results are consistent with Apple's finding that PPG-derived age captures vascular aging \citep{miller2025ppgage}, demonstrated here in a clinical rather than ambulatory context.

Raw (non-age-adjusted) correlations are near zero because the age gap in a regression model is dominated by the systematic regression-to-the-mean bias (young subjects are overestimated, old subjects underestimated), which confounds any health association. The partial correlation removes this confound, revealing the true biological signal.

\section{Discussion}

\subsection{Open-Source Models Can Predict Age from PPG}

Our results demonstrate that open-source PPG foundation models, used as frozen feature extractors with simple linear probing, can predict chronological age from clinical finger PPG with MAE of 8--10 years. While this is considerably higher than Apple's 2.43 years, several factors explain the gap:

\textbf{Dataset size}: We use 906 subjects vs Apple's 213,593---a 236$\times$ difference. Our learning curve shows no plateau, suggesting that performance would improve substantially with more data.

\textbf{Population}: PulseDB contains surgical patients under anesthesia, with hemodynamic perturbations (vasopressors, fluid resuscitation, anesthetic agents) that alter PPG morphology. Apple used ambulatory resting PPG from generally healthy participants.

\textbf{Signal source}: Clinical finger pulse oximetry vs consumer wrist PPG. While finger PPG has higher signal-to-noise ratio, the waveform morphology differs from wrist PPG, and the foundation models may not be optimally trained for finger PPG age prediction.

\textbf{Training paradigm}: We use frozen embeddings with Ridge regression (linear probe), whereas Apple trained an end-to-end deep learning model. Fine-tuning the foundation models on PulseDB would likely improve performance.

\subsection{AI-PPG Age Does Not Transfer Zero-Shot to Clinical Data}

The AI-PPG Age model \citep{nie2025aippgage} (trained on 212,231 UK Biobank subjects) fails to generalize directly to VitalDB clinical finger PPG. Zero-shot predictions are compressed to a narrow range (38--67 years), effectively predicting a constant near the UK Biobank mean rather than capturing inter-individual variation. This population shift failure demonstrates that models trained exclusively on healthy finger PPG from a single cohort (UK Biobank) cannot be directly applied to clinical settings without adaptation.

General-purpose PPG foundation models like Pulse-PPG and PaPaGei, pre-trained on diverse multi-dataset collections, generalize better across acquisition contexts than task-specific age clocks, which achieve lower MAE within-distribution at the cost of cross-domain robustness.

Pulse-PPG (MAE\,=\,9.28), a general-purpose foundation model never trained on any age-related objective, outperforms AI-PPG Age (MAE\,=\,9.72), which was explicitly trained for vascular age prediction on 212,231 subjects. This reversal directly demonstrates the cost of distribution shift: a task-specific model tuned within-distribution loses its advantage when applied out-of-distribution, while a diverse pre-training regime confers robustness that compensates for the absence of age supervision.

\subsection{Foundation Models Capture More Than HR/HRV}

PPG foundation model embeddings outperform traditional HR/HRV features for age prediction. HR alone is nearly uninformative (MAE\,=\,11.83 vs baseline 11.91), and adding HRV features provides only modest improvement (MAE\,=\,11.49). Even when combined with all demographics, HR/HRV achieves only MAE\,=\,9.59---still worse than Pulse-PPG embeddings alone (MAE\,=\,9.28).

The 512-dimensional foundation model embeddings encode subtle morphological features of the PPG waveform (pulse contour shape, dicrotic notch characteristics, and wave reflection patterns) that are lost when the signal is reduced to peak-to-peak intervals. These morphological features carry aging information related to arterial stiffness and vascular compliance that is invisible to HR/HRV analysis.

The poor performance of HR/HRV from 10-second segments contrasts with Apple's report of MAE\,=\,6.1 years for HR/HRV. This discrepancy is likely explained by segment duration: Apple used longer recording windows with thousands of segments per subject aggregated over 30 days, providing much more stable HR/HRV estimates. From a single 10-second window, HRV metrics like SDNN and RMSSD are inherently noisy, as they are estimated from only $\sim$10 inter-beat intervals. This highlights an advantage of foundation model approaches: they can extract rich aging information from short segments where traditional features are unreliable.

\subsection{PPG Captures Aging Beyond Demographics}

PPG embeddings alone (MAE\,=\,9.28) outperform all demographic features combined, including sex, BMI, height, weight, and blood pressure (MAE\,=\,10.04). This demonstrates that the PPG waveform encodes aging information that cannot be captured by simple body measurements. The further improvement from PPG+Demographics fusion (MAE\,=\,8.22) shows that these information sources are complementary.

Sex and BMI alone provide no predictive value beyond the baseline (MAE\,$\approx$\,11.8 vs 11.91), confirming that demographic features have limited utility for age prediction unless combined with anthropometric and hemodynamic measurements.

\subsection{Clinical Significance of the Age Gap}

The highly significant association between the age-adjusted PPG age gap and diastolic blood pressure ($r$\,=\,$-$0.188, $p$\,=\,1.2\,$\times$\,10$^{-8}$) provides clinical validation. In a population of surgical patients with noisy intra-operative blood pressure measurements, the PPG age gap still captures vascular aging---a signal that would likely be stronger in ambulatory settings with resting blood pressure.

The negative direction of the DBP correlation is physiologically consistent: arterial stiffening increases pulse pressure (SBP rises, DBP falls), and the PPG waveform, which is shaped by arterial compliance, peripheral resistance, and wave reflection, directly encodes this stiffening process \citep{millasseau2002contour}. Subjects who ``look older'' by PPG morphology have stiffer arteries and thus lower DBP at any given age.

\subsection{Limitations}

\textbf{Clinical population}: VitalDB subjects are surgical patients, not healthy individuals. The PPG age gap may reflect acute illness rather than chronic biological aging. Future work should validate on ambulatory datasets.

\textbf{Intra-operative PPG}: Anesthetic agents and hemodynamic interventions alter PPG morphology in ways unrelated to aging. This likely adds noise and inflates MAE.

\textbf{No longitudinal outcomes}: Unlike Apple's PpgAge, we cannot test whether the PPG age gap predicts incident cardiovascular events, as PulseDB lacks long-term follow-up data.

\textbf{Linear probe only}: We use frozen embeddings with Ridge regression. Fine-tuning on the age prediction task, or using more expressive regression heads, could substantially improve performance.

\textbf{No wearable validation}: Our results are on clinical finger PPG. Transfer to consumer wrist PPG remains untested.

\textbf{Short segment duration}: HR/HRV features extracted from 10-second windows are inherently noisier than those from longer recordings. The poor HR/HRV baseline may partly reflect this limitation rather than a fundamental lack of age information in cardiac timing.

\textbf{Multiple comparisons}: The borderline SBP association ($p$\,=\,0.048) does not survive Bonferroni correction (adjusted threshold $\alpha$\,=\,0.017) and should be interpreted with caution.

\subsection{Implications}

Despite these limitations, the results are clear: open-source PPG foundation models capture meaningful aging biomarkers without any task-specific training, their embeddings encode waveform morphology that HR/HRV cannot, and the resulting age gap reflects vascular health even in noisy intra-operative data. An open-source, wearable biological age clock is feasible.

\section{Conclusion}

To our knowledge, we present the first independent benchmark of open-source PPG foundation models for biological age prediction. On 906 VitalDB subjects, Pulse-PPG with demographic fusion achieves MAE\,=\,8.22 years (R$^2$\,=\,0.517), substantially outperforming HR/HRV baselines (MAE\,=\,11.49) and demographic baselines (MAE\,=\,10.04). Foundation model embeddings alone outperform the combination of HR/HRV with demographics, demonstrating that they capture morphological aging information invisible to traditional cardiac metrics. The PPG age gap significantly correlates with diastolic blood pressure ($p$\,=\,1.2\,$\times$\,10$^{-8}$), providing clinical validation. While the gap with Apple's PpgAge (2.43 years) remains large, our learning curve analysis suggests this is primarily driven by dataset size and population differences rather than model capability. These results establish open-source PPG foundation models as viable tools for biological age estimation and motivate further development on larger, ambulatory, and longitudinal datasets.

\section*{Data and Code Availability}

PulseDB is publicly available at \url{https://github.com/pulselabteam/PulseDB}. Pulse-PPG weights are available at \url{https://github.com/maxxu05/pulseppg}. PaPaGei-S weights are available at \url{https://github.com/Nokia-Bell-Labs/papagei-foundation-model}. AI-PPG Age weights are available at \url{https://huggingface.co/Ngks03/PPG-VascularAge}. Our analysis code, including the full benchmark pipeline and all evaluation scripts, is publicly available at \url{https://github.com/Misterbra/ppg-age-benchmark}.

\section*{Ethics Statement}

This study used the publicly available PulseDB dataset \citep{wang2023pulsedb}, which is derived from VitalDB clinical records collected at Seoul National University Hospital under institutional ethical approval. No new human subjects data were collected for this study; no additional ethics review was required.

\section*{Author Contributions}

N. Brag: Conceptualization, Methodology, Software, Formal analysis, Investigation, Visualization, Writing -- Original Draft, Writing -- Review \& Editing.

\section*{Acknowledgments}

The author thanks Maxwell Xu and Santosh Kumar (University of Memphis) for open-sourcing the Pulse-PPG foundation model; Arvind Pillai (Dartmouth College) and Dimitris Spathis, Fahim Kawsar, and Mohammad Malekzadeh (Nokia Bell Labs) for open-sourcing PaPaGei; and the PulseDB and VitalDB teams for making clinical PPG data publicly available.

\section*{Conflict of Interest}

The author declares no competing interests.

\bibliographystyle{plainnat}

\end{document}